Short Paper*

# Smart Metro: Deep Learning Approaches to Forecasting the MRT Line 3 Ridership

Jayrald Empino
Department of Computer Science, Technological University of the Philippines,
Manila, Philippines
jayrald.empino@tup.edu.ph

Jean Allyson Junsay
Department of Computer Science, Technological University of the Philippines,
Manila, Philippines
jeanallyson.junsay@tup.edu.ph

Mary Grace Verzon
Department of Computer Science, Technological University of the Philippines,
Manila, Philippines
marygrace.verzon@tup.edu.ph

Mideth Abisado
College of Computing and Information Technologies, National University,
Manila, Philippines
mbabisado@national-u.edu.ph

Shekinah Lor Huyo-a
Research and Development Center, Philippine Coding Camp,
Manila, Philippines
shekinah@philippinecoding.com

Gabriel Avelino Sampedro
Faculty of Information and Communication Studies,
University of the Philippines Open University, Laguna, Philippines
garsampedro@ieee.org
(corresponding author)

**Abstract**

*Purpose* – Since its establishment in 1999, the Metro Rail Transit Line 3 (MRT3) has served as a transportation option for numerous passengers in Metro Manila, Philippines. The Philippine government's transportation department records more than a thousand people using the MRT3 daily and forecasting the daily passenger count may be rather challenging. The MRT3's daily ridership fluctuates owing to variables such as holidays, working days, and other unexpected issues. Commuters do not know how many other commuters are on their route on a given day, which may hinder their ability to plan an efficient itinerary. Currently, the DOTr depends on spreadsheets containing historical data, which might be challenging to examine. This study presents a time series prediction of daily traffic to anticipate future attendance at a particular station on specific days.

*Method* – The proposed prediction approach uses DOTr ridership data to train multiple models that can provide correct data on Azure AutoML. These trained models have the highest accuracy: Gradient Boosting, Extreme Random Trees, and Light GBM.

*Results* – Based on historical data, this study aims to build and evaluate several prediction models for estimating the number of riders per station. On Azure AutoML, the Gradient Boosting, Extreme Random Trees, and Light GBM algorithms were investigated and executed. Gradient Boosting and Extreme Random Trees frequently made the most accurate predictions of the three algorithms, with an average accuracy of over 90%.

*Conclusion* – This research aims to develop and test different models of prediction for forecasting the number of riders per station based on historical data. Seven days of data were utilized for applying the model or assessing its correctness. Each model's resultant accuracy in each station is unique and may be modified by ridership and geography. However, the model still provides complete precision. Accuracy may be enhanced if additional current, valuable, and efficient characteristics are introduced to the dataset.

MRT3 might incorporate a mortality rate component into the station's relative location or passenger capacity.

*Recommendation* – As the acquired data were from a pandemic, it is suggested that additional information be employed in future research. The circumstances of the MRT might change substantially over time; therefore, it is essential to refresh the training dataset.

*Practical Implication* – There are several benefits to applying time series forecasting in predicting the ridership of the MRT3 in the Philippines. This can allow decision-makers to make informed decisions about optimizing the MRT3 system to meet the needs of commuters. Additionally, time series forecasting can help to identify potential problems or issues in advance, such as overcrowding or maintenance needs, allowing for proactive solutions to be implemented.

*Keywords* – Light GBM, Gradient Boosting, Extreme Random Trees, Time Series Forecasting

## INTRODUCTION

The MRT3, also known as the Metro Rail Transit Line 3, is a rapid transit system in the Philippines. It runs along EDSA, a major thoroughfare in Metro Manila, and serves approximately 7% of the total population of the Philippines (Philippine Statistics Authority, 2021). The MRT3 was built in the late 1990s and early 2000s to alleviate traffic congestion in the city and provide a more efficient public transportation option for commuters. It is operated by the Metro Rail Transit Corporation and serves millions of passengers annually. The MRT3 has become an integral part of the transportation infrastructure in Metro Manila and is an essential part of daily life for many residents. Unfortunately, the MRT3 is often overcrowded, with long lines and packed trains becoming common. 70% of passengers in Metro Manila rely on public transport, and commuters have complained about the difficulty of finding a seat or even standing room on the trains, leading to a frustrating and uncomfortable travel experience (Ito, 2022). Despite efforts to improve capacity and efficiency, the MRT3 remains a crowded and congested transportation option for many Filipinos. This research will focus on the MRT3, which links to one of the country's premier central business areas, Makati.

The MRT3 operates on a single line stretching 13.8 kilometers from the North Avenue Station in Quezon City to the Taft Avenue Station in Pasay City. It has 13 stations along its route, including notable stops at EDSA, Ayala, and Makati. The trains run on a standard gauge track and are powered by an overhead catenary system. Each train consists of four cars, with a total capacity of up to 1,000 passengers. The MRT3 has a maximum speed of 80 kilometers per hour. However, it typically operates at a slower speed due to the high volume of passengers and the densely populated areas it serves. The system also has a communication-based train control system, which helps improve the trains' reliability and safety. Despite these technical features, the MRT3 has faced numerous challenges recently, including issues with its aging infrastructure, maintenance problems, and overcrowding (Dalmacio, et al., 2019).

Transportation in the Philippines has been significantly impacted by the COVID-19 pandemic (Chuenyindee, et al., 2022). To curb the spread of the virus, the government implemented various measures, including the suspension of public transportation and strict health protocols. Despite this, the International Association of Public Transportation stresses the need for public transportation to promote mobility during the crisis (Des Transport Publics, 2020). This has significantly impacted commuters, particularly those who rely on public transportation to get to work or school. In response, the government has implemented alternative transportation options, such as shuttle services and bike-sharing programs. However, these options are often limited and must thoroughly address all commuters' needs. In addition, implementing health protocols, such as the requirement for face masks and the reduction of capacity on public transportation, has further exacerbated the already crowded conditions on many modes of transportation. Overall, the COVID-19 epidemic has disrupted transportation in the Philippines and presented substantial obstacles to commuters since many Filipinos simultaneously continue to go to work, school, and home in Metro Manila. The roads become highly crowded with public and private cars; hence, the MRT replaces buses and jeepneys (Guno, et al., 2021). Consequently, the MRT might become busy at certain hours of the day. Thousands of commuters have difficulty fitting onto MRT vehicles, and the number of passengers regularly surpasses the MRT's optimal capacity.

This study intends to produce a data-driven forecast of MRT3 ridership at particular stations and at certain times of the day. The suggested model must incorporate Light GBM, Gradient Boosting, and Extreme Random Trees to provide reliable predictions. By offering a data-driven forecast of the volume of passengers at a particular moment, commuters may plan for potential rush hours and explore alternative ways of getting around.

There are several benefits to applying time series forecasting in predicting the ridership of the MRT3 in the Philippines. First, time series forecasting can provide a reliable estimate of future ridership based on past data, which can be used to plan for resources and infrastructure needs. By analyzing patterns and trends in the data, time series forecasting can help to identify factors that may influence ridership, such as the time of day, weather conditions, or holidays. This can allow decision-makers to make informed decisions about optimizing the MRT3 system to meet the needs of commuters. Additionally, time series forecasting can help to identify potential problems or issues in advance, such as overcrowding or maintenance needs, allowing for proactive solutions to be implemented. Overall, time series forecasting can provide valuable insights and support decision-making for improving the efficiency and effectiveness of the MRT3 system.

This paper is structured as follows: Section II summarizes relevant studies and efforts on machine learning approaches to ridership forecasting. The third section gives

1926

background information on the algorithm and approach utilized to construct the current study. In Section IV, the findings of the experiment are presented. Section V concludes with a summary and recommendations for further study.

**LITERATURE REVIEW**

Time series forecasting is a method of predicting future values based on past data points collected at regular intervals over some time. It is commonly used in various industries to make informed decisions about future events, including finance, economics, and meteorology. Time series forecasting uses statistical models to analyze patterns and trends in the data and predict future values. These models can be either qualitative, based on expert knowledge and subjective judgment, or quantitative, based on statistical data analysis. Time series forecasting can be used to make short-term predictions, such as daily or hourly forecasts, or long-term predictions, such as yearly or decade-long forecasts. It is an essential tool for businesses and organizations to plan for the future and make informed decisions about resource allocation and strategy. Many algorithms can be used to train and predict time series data, such as ARIMA, prophet, and SARIMA; all these algorithms are used for specific use cases, like if there are trends, seasonality, and irregularities in the data. This section shall discuss some of the advances in the field of ridership prediction.

Shen et al. describe creating a model that forecasts short-term subway ridership by employing a gradient-boosting decision tree (GBDT) to train and assess the model's performance (Shen, et al., 2020). The accuracy is between 0.98 and 0.99, indicating that the model provides promising results. Gradient boosting achieved better outcomes by giving the residuals required to construct the decision tree at each iteration. As a result, the results demonstrated that GBDT greatly surpasses other techniques in prediction accuracy and model interpretation ability, resulting in anticipated short-term subway ridership with high and objective accuracy.

A paper by X. Ma, et al. (2018) presents a machine learning model for predicting ridership in metro systems. The model combines convolutional neural networks (CNNs) and long short-term memory (LSTM) networks in a parallel architecture to analyze data from multiple sources, including passenger flow, weather, and economic indicators. The model is tested on real-world data from the Beijing subway system and is found to have high accuracy in predicting ridership. The authors argue that the model can be applied to other metro systems and could be useful for optimizing resource allocation and improving efficiency.

Another paper by Chen, et al. (2020) presents a machine learning model for predicting short-term metro ridership. The model combines two techniques: seasonal and trend decomposition using loess (STL) and long short-term memory (LSTM) neural networks. The STL technique identifies and removes trends and seasonal patterns from the data, while the LSTM networks are used to analyze the remaining data and make predictions. The model is tested on real-world data from the metro system in China and is found to have high accuracy in predicting ridership. The authors argue that the model can be applied to other metro systems and could be useful for optimizing resource allocation and improving efficiency.

Using the elastic net and fuzzy cognitive mappings (FCMs), Liu et al. (2020) overcome the difficulty of processing multivariate long stationary timer series, such as electroencephalogram (EEG) data. In the study, human activities were predicted using EEG data by first predicting EEG patterns using historical data. The grasp-and-lift dataset demonstrated that the model could predict EEG data with a decreased prediction error. Consequently, the researchers concluded that multivariate long stationary time series might be accurately predicted and identified by employing elastic nets and fuzzy cognitive maps.

A paper by Lv et al. (2021) focuses on predicting stock prices through a machine learning approach. Long short-term memory (LSTM) was employed in the study, but due to the inefficiency of the model, the researchers opted for LightGBM-Optimized LSTM. Comparative models, especially RNN and GRU, are also utilized in this work. The testing results demonstrate that LightGBM-Optimized LSTM continues to function optimally, with the highest prediction accuracy in following stock index price movements. Therefore, the researchers determined that only LightGBM-Optimized LSTM is near the actual value for all three (3) models.

## METHODOLOGY

This study's objective is to investigate the use of deep learning models to forecast MRT3 ridership patterns. Three distinct machine learning models must be employed: LightGBM, Gradient Boosting, and Extreme Random Trees. In this part, the various construction methods of the proposed system will be detailed. The phases of data collection, data cleansing, and machine learning model development are covered.

### *Data Gathering/Collection*

At this point, it is assumed that the data will have the necessary elements, such as timestamp and ridership; there are sufficient features to train the specified model. The dataset to be utilized is obtained through a freedom of information (FOI) request to the transportation department. The dataset contains the daily ridership of the various MRT3 stations for 2019 through 2021. The data received included every station's timestamp, point-of-entry, point-of-exit, total entry, and exit.

### *Data Cleaning*

In this step, irrelevant data elements will be eliminated, and useful ones will be introduced. The timestamp, entrance, and station remain. Furthermore, the data will be augmented with two components: weekends and holidays. Therefore, the dataset will have four (4) columns: timestamp, ridership (entry), weekend, and holiday. Weekends and holidays shall have a binary value of either 1 or 0. If it is a weekend or holiday, it is noted with 1; otherwise, it is marked with 0. In addition to irrelevant data, there were duplicates among the data for the first two months of the year. The number of monthly passengers at the North Avenue station of the MRT3 is depicted in Figure 1 based on data collected in 2020.

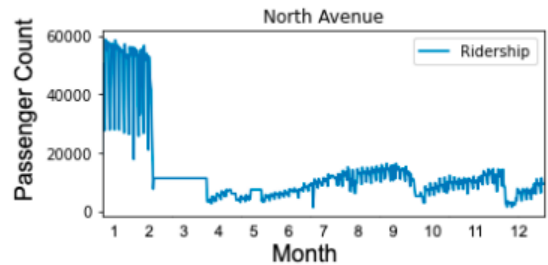

*Figure 1.* The graph plots the data from the 2020 ridership for the MRT3 North Avenue Station.

The data are cleansed to eliminate discrepancies and duplications, and Fig. 2 depicts the ridership for 2020.

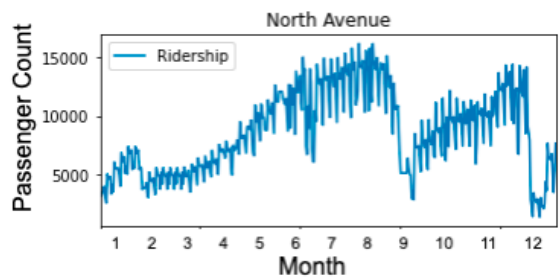

*Figure 2.* The cleaned data of North Avenue MRT3 station shows the actual number of passengers per month for 2020.

While Figure 2 displays data from the North Avenue station, data from other stations were also collected. Due to abrupt lockdowns, changes in covid cases, and technological challenges, the data was forced to drop, as seen in the charts above; it increases and decreases cyclically. The data exhibits a trend, but the noise impacts the time series. Therefore, it is essential to understand the noise and distinguish it from the actual trend.

### *Data Modeling*

In modeling the data, Azure AutoML is used to train different models. Azure AutoML is a machine learning platform offered by Microsoft that allows users to create and train predictive models without the need for coding or technical expertise. It is designed to simplify the process of building and deploying machine learning models by providing a range of automated features and tools. With Azure AutoML, users can easily access a wide range of pre-built models and algorithms and customize their models using a simple interface. The platform also offers a range of features to improve the accuracy and performance of the models, including data preprocessing, feature engineering, and model selection. Azure AutoML also provides real-time performance monitoring and the ability to deploy the models to various environments, such as Azure Machine Learning, Azure Kubernetes Service, or Azure Functions. Overall, Azure AutoML is a powerful tool for organizations looking to leverage the power of machine learning in their operations and make data-driven decisions. Azure AutoML trains and validates models automatically using k-fold cross-validation in this study. Each station uses its model, which may be the same as other stations based on the best-trained AutoML model.

K-fold cross-validation is a commonly used technique in machine learning for evaluating the performance of a model. It involves dividing the available data into a specified number of "folds," or subsets, and training the model on a different subset each time while evaluating its performance on the remaining data. This process is repeated until the model has been trained and evaluated on all the folds. The final performance score is calculated as the average performance across all folds. Fig. 3 illustrates how the data was split into two (2) train and test data. The train data was used in the first part of the dataset to help build the prediction model. As for the test data, this is used to validate the model and further enhance accuracy. The training data is rolling and used against the new test data in every fold. Hence, the proponents used five folds for the cross-validation.

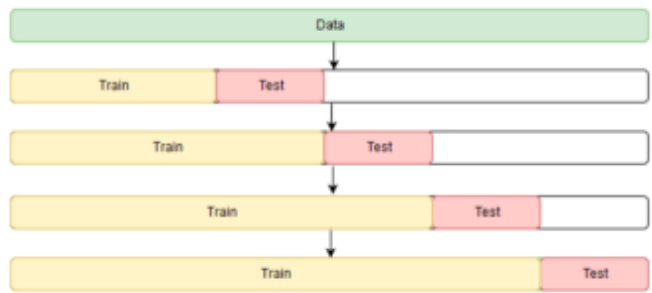

*Figure 3.* Cross Fold Validation in Rolling Basis

To train the models, the AutoML run must be built first. The following configuration requirements must be met:

1) Required dataset - Each station's dataset has been uploaded to the Azure Dataset Store.
2) Experiment container - To train the models, it is necessary first to build an experimental container that will be utilized for associated AutoML runs. Proponents designated it "mrt3."

3) The target column for prediction must specify which column is to be predicted so that AutoML is aware of the machine learning type to employ. It is based on the ridership in the dataset.
4) Virtual machine – AutoML will utilize this computational environment to train the models.
5) Machine learning type – AutoML may pick it automatically depending on the dataset and column to forecast. However, it may still be specified manually. In the instance of a project, time forecasting was utilized.
6) Time column - AutoML must be aware of which column contains the time so that it may readily get it. The advocates employ the "timestamp."
7) Data frequency (hourly, daily, or weekly) — The dataset contains historical data; as the project focuses on daily ridership, the daily frequency should be selected.
8) Forecast horizon - This is when a model can accurately anticipate the future. The proponents set it at 7, corresponding to one week or seven days, as the dataset needs to be more comprehensive to forecast higher than that.

The model will train continuously for a maximum of 6 hours, and it will do iteration for each model and compare it with other models to determine if it improves — the major measure used is the normalized root mean squared error. AutoML will run continually until the model has not improved after 20 iterations. Each station's dataset was subjected to training for about thirty minutes to identify a viable model for use in forecasting.

## RESULTS AND DISCUSSION

After all experiments and AutoML configuration runs, the models are complete. Each station trains a unique number of models. By default, AutoML returns the voting ensemble as the best model; it mixes many models to predict more accurate outcomes. The voting ensemble is ruled out of scope, and the second-best model will be prioritized.

Mean absolute percentage error (MAPE) is present in the first findings; it will be deleted and replaced with its opposite correctness. Also included in the table is the algorithm and construction time. Specific stations utilize the first algorithm, which has the highest precision.

The graph below depicts the aggregated feature significance of the input dataset used by the first algorithm listed in the table. It indicates which columns contribute significantly to total correctness. All models have the same feature significance, but their aggregated value varies.

The preceding charts and tables show that all stations have the same feature weight. Timestamps are a significant factor in improving precision. While the weekend and holiday

– the two additional characteristics added to the dataset – also aid the algorithm in predicting actual values as accurately as possible.

Azure AutoML trained several models and completed at least three iterations. Gradient Boosting and Extreme Random Trees frequently achieved the highest trained model precision. The table below displays the number of stations that used the specified model.

Table 1. Number of stations used the model

| Algorithm | Number of Station |
|---|---|
| Gradient Boosting (GB) | 7 |
| Extreme Random Trees (ERT) | 5 |
| Light GBM (LGBM) | 1 |

Based on Table I, gradient boosting has a more significant number of stations than extreme random trees, which also have a more significant number of stations. One station only uses the light GBM model. The trained model has a prediction horizon of 1, indicating that it was trained to anticipate ridership per day. The prediction input data spans seven days, from September 1 to September 7, 2021.

Table 2. Predicted Accuracy for All Stations

| Station | Algorithm | Accuracy |
|---|---|---|
| North Avenue | GB | 86.18% |
| Quezon Avenue | ERT | 88.69% |
| GMA Kamuning | ERT | 77.4% |
| Ayala Avenue | GB | 96.02% |
| Bonifacio Avenue | GB | 96.97% |
| Buendia | ERT | 89.96% |
| Cubao | GB | 95.39% |
| Guadalupe | LGBM | 83.29% |
| Magallanes | GB | 95.17% |
| Ortigas | ERT | 83.75% |
| Santolan | GB | 84.49% |
| Shaw Blvd | GB | 95.18% |
| Taft Avenue | ERT | 90.53% |

## CONCLUSION AND RECOMMENDATION

Based on historical data, this study aims to build and evaluate several prediction models for estimating the number of riders per station. On Azure AutoML, the Gradient Boosting, Extreme Random Trees, and Light GBM algorithms were investigated and executed. Gradient Boosting and Extreme Random Trees frequently made the most accurate predictions of the three algorithms. Each station has its optimum predictive model for ridership. Mean Absolute Percentage Error (MAPE) or the reverse, which is accuracy, is used to compare the accuracy of the two best-trained models. Only the first best model is utilized for prediction implementation. Timestamp, weekend, and holiday have the same relevance across all stations; only the aggregated values of each attribute vary. The timestamp is the most important information since it is the foundation for the model predicting increased and decreased riding periods.

Seven days of data were utilized for applying the model or assessing its correctness. Each model's resultant accuracy in each station is unique and may be modified by ridership and geography. However, the model still provides complete precision. Accuracy may be enhanced if additional current, valuable, and efficient characteristics are introduced to the dataset. MRT3 might incorporate a mortality rate component into the station's relative location or passenger capacity.

As the acquired data were from a pandemic, it is suggested that additional information be employed in future research. The circumstances of the MRT might change substantially over time; therefore, it is essential to refresh the training dataset.

## DECLARATIONS

### Conflict of Interest

The author declares that there is no conflict of interest.

### Informed Consent

Not applicable, but all requirements to acquire the datasets are registered through eFOI. In addition, the datasets are considered public data.

### Ethics Approval

Not applicable. No human or animal subjects were involved in this study.

**Author's Biography**

Jayrald Empino is an undergraduate researcher in the field of deep learning, with a focus on developing and improving models for various applications. Hi is currently affiliated with the Department of Computer Science at the Technological University of the Philippines in Manila. His research interests include machine learning, computer vision, and natural language processing.

Jean Allyson Junsay is an undergraduate researcher in the field of deep learning, with a focus on developing and improving models for various applications. She is currently affiliated with the Department of Computer Science at the Technological University of the Philippines in Manila. Her research interests include machine learning, computer vision, and natural language processing.

Mary Grace Verzon is an undergraduate researcher in the field of deep learning, with a focus on developing and improving models for various applications. She is currently affiliated with the Department of Computer Science at the Technological University of the Philippines in Manila. Her research interests include machine learning, computer vision, and natural language processing.

Dr. Mideth Abisado is an Associate Member of the National Research Council of the Philippines and a Board Member of the Computing Society of the Philippines Special Interest Group for Women in Computing. She is the Director of the CCIT Graduate Programs Department of the National University, Manila. She heads research on Harnessing Natural Language Processing for Community Participation. Social science, machine learning, and natural language processing principles and techniques are used in the study. It is well anticipated that thematic based on dashboard analytics will be used for policy recommendations for the government. Her research focuses on Emphatic Computing, Social Computing, Human-Computer Interaction, and Human Language Technology. She has 23 years of experience in education and research. Her passion is to encourage women to choose careers in computing and prepare and mold the next generation of Filipino IT professionals and leaders in the country.

Shekinah Lor Huyo-a is a researcher at the Research and Development Center of Philippine Coding Camp in Manila, Philippines, who is passionate about developing innovative solutions using cutting-edge technology. Her research areas include deep learning techniques in various fields, specifically natural language processing, computer vision, and data analysis.

Gabriel Sampedro received his M.S. degree in Computer Engineering from Mapua University. He started working as a firmware engineer at an engineering design firm in 2018 before shifting his focus to the startup industry as an entrepreneur and in the academe as an assistant professor. He is a strong advocate of technology education, as he believes in the unrealized potential of the Philippines in the tech industry. Through his passion, he helped build different tech startups like Philippine Coding Camp, Inc. (Manila Coding Camp) and MachiBox Inc. In addition to managing his startups, is currently taking up his Ph.D. in I.T. Convergence Engineering as a doctorate researcher at Kumoh National Institute of Technology (South Korea); and an Assistant Professor 2 at the University of the Philippines - Open University.